\documentclass[runningheads]{llncs}

\usepackage[T1]{fontenc}
\usepackage{subfigure,graphicx}
\usepackage{amsmath,amsfonts,latexsym,amssymb,euscript}
\usepackage{booktabs}
\usepackage[nodayofweek]{datetime}
\usepackage{hyperref}
\usepackage[table]{xcolor}
\usepackage{color,colortbl,tabularx}
\usepackage[english]{babel}
\usepackage{amsfonts}
\usepackage{pifont}
\usepackage[inline]{enumitem}

\usepackage{tikz,tikz-cd}
\usetikzlibrary{arrows,cd,positioning}
\usetikzlibrary{shapes,snakes}

\usepackage{boxedminipage}

\newif\iflong
\newif\ifshort

\longtrue

\iflong
\else
\shorttrue
\fi

\usepackage[noend]{algpseudocode}
\usepackage{algorithm,algorithmicx}

\algnewcommand\algorithmicinput{\textbf{Input:}}
\algnewcommand\INPUT{\item[\algorithmicinput]}

\algnewcommand\algorithmicoutput{\textbf{Output:}}
\algnewcommand\OUTPUT{\item[\algorithmicoutput]}

\newcommand{\DI}{\mathit{Di}}
\newcommand{\PA}{\mathit{Pa}}
\newcommand{\MU}{\mathit{Mu}}
\newcommand{\DR}{\mathit{Dr}}

\newcommand{\mcup}[1]{\MU_{\cup}(#1)}
\newcommand{\mcap}[1]{\MU_{\cap}(#1)}

\newcommand{\rest}[2]{#1_{|#2}}

\begin{document}
\title{Toward a Unified Graph-Based Representation of Medical Data for Precision Oncology Medicine}
\titlerunning{Graph-Based Representation for Precision Oncology Medicine}

\author{Davide Belluomo\inst{1}\and
Tiziana Calamoneri\inst{1}\orcidID{0000-0002-4099-1836} \and
Giacomo Paesani\inst{1}\orcidID{0000-0002-2383-1339} \and
Ivano Salvo\inst{1}\orcidID{0000-0003-3111-701X}}

\authorrunning{D. Belluomo et al.}

\institute{$^1$Computer Science Department, Sapienza University of Rome
\email{dav.belluomo@gmail.com}\\
\email{\{calamo,paesani,salvo\}@di.uniroma1.it}}

\maketitle

\begin{abstract}
We present a new unified graph-based representation of medical data, combining genetic information and medical records of patients with medical knowledge via a unique knowledge graph. This approach allows us to infer meaningful information and explanations that would be unavailable by looking at each data set separately. The systematic use of different databases, managed throughout the built knowledge graph, gives new insights toward a better understanding of oncology medicine. Indeed, we reduce some useful medical tasks to well-known problems in theoretical computer science for which efficient algorithms exist.

\end{abstract}

\section{\bf Introduction}\label{sec:SCIENTIFIC-BACKGROUND}\vspace{-2mm}

One of the recent and numerous applications of graph theory is {\it network medi\-cine}~\cite{BGL11} that aims to identify, prevent, and treat diseases: graph-based approaches have offered an effective tool to systematically explore the intrinsic complexity of diseases, leading to the identification of disease specificity, disease-associated genetic mutations and a new way to assign treatments to patients. A new approach in the framework of network medicine is the so-called {\it personalized} or {\it precision} medicine, that is, the systematic use of individual patient characteristics to determine which treatment option is most likely to result in a better average outcome for the patient.

In particular, a new trend in precision oncology aims to shape drug treatments based on the specific gene mutation profile of the particular patient (see, for example,~\cite{LW99}). In the last 20 years, medical practitioners tested this new approach and obtained an improvement regarding its effectiveness. However, precision medicine developments have not been in line with the expectations. A common belief among oncologists and researchers in bioinformatics is that the discrepancy between real and expected performance can be partially explained by looking at the role of gene mutations in cancer evolution: the ambitious long-term goal of our research is to contribute to filling the gap in medical knowledge by examining and comparing genetic profiles of an extensive database of patients together with different treatment outcomes.

In computer science, graphs and networks are widely exploited data structures to store information as they can represent complex systems as sets of binary interactions or relations between various entities: in particular, it is possible to encode the information stored in different databases in a single graph. 

Our research embraces this line of research: exploiting various databases at the same time, we represent all the available data regarding genetic information and medical records of a group of patients, together with medical knowledge in a unique {\it knowledge graph} and perform a guided analysis of some medical issues. 

In particular, in Section~\ref{sec:DATA-AND-METHODS}, after giving some preliminary definitions, we formally describe how we construct our knowledge graph. Then, we show how some medical issues can be modeled as graph problems and solved through classical graph algorithms. Section~\ref{sec:RESULTS} is devoted to showing the results of some preliminary experiments as a proof of concept. Finally, Section~\ref{sec:CONCLUSIONS} concludes the paper.
\vspace{-8mm}

\section{\bf Data and Methods}\label{sec:DATA-AND-METHODS}

\vspace{-2mm}
The main idea of this work is to collect together as much information as possible, coming from the structured data of different databases recording data from medical studies, and official documents of regulatory agencies, in order to study oncological diseases and, in particular, relationships among gene mutations, diseases, and treatment effectiveness, and try to infer vital information supporting the medical community. 

In the following, we propose to use a graph (the knowledge graph $H$ described in Subsection~\ref{sec:knowledgeGraphH}) to represent all such information in a uniform way. This approach has several advantages: we can give support to study and reduce medical issues by means of well-studied graph problems that, in turn, have well-known solutions based on efficient graph algorithms. Moreover, we can exploit the flexibility of graphs as a data structure, and easily support a function for quickly updating the information stored in the graph $H$; this is especially useful when new medical studies are published or when a new drug is individuated, and it is useful to incorporate this information in the graph. It is worth noting that this approach is in contrast to the static graphs created after a training phase and using them as predictive models. 
\vspace{-5mm}

\subsection{Preliminary Definitions}\label{sec:PRELIMINARY-DEF}

\vspace{-2mm}
We choose to favor intuition over formalism so, in this subsection, we informally give some basic definitions concerning graphs that will be useful in the following. The reader interested in a more formal setting can refer, for example, to~\cite{diest}.

A {\it graph} $G=(V, E)$ is constituted by a finite set $V$ of elements, known as {\it nodes}, and a collection of {\it edges} $E$, connecting pairs of nodes, representing some kind of binary relation. It is possible to label some nodes and/or edges to annotate them with additional information. The set of nodes that are connected through an edge to the same node $v$ constitute the {\it neighborhood} of $v$, and a {\it path} of $G$ is a sequence of edges that joins a sequence of nodes.

A graph $G'=(V',E')$ is a {\it subgraph} of $G=(V,E)$ if $V'\subseteq V$, $E'\subseteq E$, and every edge in $E'$ has both endpoints in $V'$. The {\it subgraph induced by a node set} $U\subseteq V$ is the graph whose nodes are all the nodes in $U$ and whose edges are all the edges present in $G$ such that both endpoints are in $U$.

A graph $G=(V, E)$ is {\it $k$-partite} if its node set $V$ can be partitioned into $k>1$ subsets, and no edge connects two nodes from the same subset. A $2$-partite graph is also known as {\it bipartite}.
\vspace{-5mm}

\subsection{The Knowledge Graph $H$}\label{sec:knowledgeGraphH}

\vspace{-1mm}
The graph $H$, which is the core of this work, is built as the union of three graphs: the graph $G$, storing {\it Genetic information} of a group of patients (whose edges are colored in Green); the graph $R$, storing information from patient {\it medical Records} (whose edges are colored in Red), and  the graph $M$, storing some general information that we will call {\it Medical knowledge} (whose edges are colored in Magenta). 

While graphs $G$, $R$, and $M$ share some (set of) nodes (for example, the set of patients), their edge sets are pairwise disjoint. 

\smallskip
\noindent 
The green graph is bipartite and is defined as $G=(\PA\cup\MU, E_G)$ where:

\begin{itemize}[wide, itemsep=0.5mm, topsep=1mm]
\item $\PA$ is a set of encrypted recorded {\it patients} in the database; a mapping $\rho~:~\PA\rightarrow \mathbb{N}$ labels every patient $p\in \PA$ with their {\it survival period}, that is the time interval that spans between the diagnosis of a specific disease and the time of the study, if the patient is still alive, or the time of the patient's death, otherwise. Patients are also labeled with a boolean mapping $\alpha~:~\PA\rightarrow \{T, F\}$ that represents whether the patient is alive or not at the time of the study: for every $p\in \PA$, $\alpha(p)=T$ if $p$ is alive and $F$, otherwise.
\item $\MU$ is a set of {\it gene mutations}. Observe that a gene can have more than one mutation; 
\item the set of green edges $E_G$ contains an edge $(p, m)$ if the patient $p$ has the mutation $m$, and this edge is labeled with the variant allele frequency (VAF) associated with $m$ for patient $p$;
\end{itemize}

\smallskip
\noindent 
The red graph is 3-partite and defined as $R=(\PA\cup\DI\cup\DR, E_R)$ where:

\begin{itemize}[wide, itemsep=0.5mm, topsep=1mm]
\item $\PA$ is the set of patients as defined in graph $G$;
\item $\DI$ is a set of {\it oncological diseases}, that affect patients in $\PA$; 
\item$\DR$ is a set of {\it drugs} of interest, possibly labeled with a string $\beta$ annotating adverse effects; 
\item the set of red edges $E_R$ contains:
\begin{itemize*}[itemsep=1mm, topsep=1mm]
\item an edge $(d, p)$ if the patient $p$ is affected by the disease $d$;
\item an edge $(p, f)$ if the patient $p$ has been treated with the drug $f$; these edges are labeled with a pair of values $(t,e)$ where $t \in \mathbb{N}$ is the number of treatments preceded it and $e \in \{p,u,r,n\}$ (standing for positive, unaltered, reduced, negative) represents the effectiveness of that drug; clearly, more drugs could have the same value of $t$ when a cocktail of drugs has been administrated;
\end{itemize*}
\end{itemize}

\smallskip
\noindent
The magenta graph is 3-partite and defined as $M=(\MU\cup\DI\cup\DR, E_M)$ where:

\begin{itemize}[wide, itemsep=1mm, topsep=0mm]
\item  $\MU$, $\DI$, and $\DR$ are defined as in graphs $G$ and $R$;
\item the set of magenta edges $E_M$ contains:
\begin{itemize*}[itemsep=1mm, topsep=1mm]
\item an edge $(d, m)$ if it is known that typically the disease $d$ appears in presence of the gene mutation $m$; these edges can be labeled with a measure that quantifies the relevance of $m$ with respect to $d$.
\item an edge $(m, f)$ if it is known that the drug $f$ has some effect on the mutation $m$.
\end{itemize*}
\end{itemize}

\begin{figure}[ht]
\centering
\vspace*{-7mm}
\includegraphics[scale=.6]{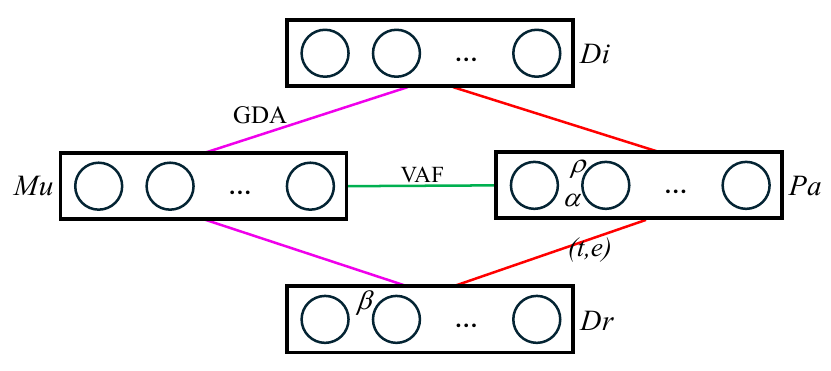}
\vspace*{-4mm}
\caption{A schematization of the knowledge graph $H$: green lines represent edges of $G$ (genetic information), red lines represent edges of $R$ (information from medical records) and magenta lines represent edges of $M$ (medical knowledge). $Di$ is the set of nodes representing the diseases, $\PA$ the patients, $\MU$ the genetic mutations, and $\DR$ the drugs.}
\label{fig.grafo}
\vspace{-12mm}
\end{figure}

\smallskip
The graph $H=(V,E)$ (see Figure~\ref{fig.grafo}) is the union of its subgraphs $G$, $R$ and $M$. Therefore the set of nodes is $V=\PA \cup \MU \cup \DI \cup \DR$, and the set of edges is $E = E_G\cup E_R\cup E_M$, that is, the union of green, red, and magenta edges as defined above. Note that no edge of $H$ has both endpoints in the same node subset; hence, $H$ is $4$-partite.

\subsection{Experiment Design}\label{sec:expermimentDesign}

\vspace{-2mm}
In our experiments, we constructed the green graph $G$ by exploiting database {\it cBioPortal}~\cite{de2023analysis,cerami2012cbio,gao2013integrative}, from which we extract the genetic information of each patient. Gene mutations are derived from NGS (Next Generation Sequencing).

The information encoded in the red graph $R$ can be obtained from the examination of a large number of medical records, which not only keeps track of the diseases affecting the patients, but also the treatment history, a quantitative estimation of the responses, and the survival period. In our experiments, we use cBioPortal again, to deduce this information.

At least in principle, the magenta graph $M$ should store the vast and ever-evolving medical knowledge. In particular, in our experiments, we consider the {\it DisGeNET} database~\cite{pinero2020disgenet} and regulatory agencies databases  (see, for example, {\it PharmGKB} \cite{CHGSTWK21,CMHGSTAK12}) for the information about the interaction between specific diseases and gene mutations, and for approved drugs targeting gene mutations, respectively.

Through knowledge graph $H$, we can answer many questions by exploiting graph algorithms. The underlying idea is that the involvement of multiple databases and subgraphs gives us robust and precise machinery to deduce meaningful information. Here, for each of the three listed examples, we describe the medical issue, we model it into a graph problem, propose an algorithmic solution, and highlight which part of the knowledge graph $H$ is involved.
\vspace{-4mm}

\subsubsection{Comparing medical knowledge and data evidence.}\label{Sec_validate_medical_knowledge}

A natural issue is understanding whether the medical knowledge agrees with information that can be inferred from medical records. Using the information in the graph $H$, a possible question is the following: does data evidence from $G \cup R$ match the medical knowledge stored in $M$? An answer to this question can improve the understanding of the relationship between diseases and gene mutations.

On one hand, thanks to the red edges of $H$, fixing a disease $d$, we consider set $\PA(d)$ of all patients affected by $d$, corresponding to all the nodes in the neighborhood of $d$ in $\PA$.  We define $\mcup{d}$ as the set of all gene mutations affecting at least one patient in $\PA(d)$ and $\mcap{d}$ as the set of all mutations affecting every patient in $\PA(d)$ (through green edges).

On the other hand, we determine through magenta edges the set $\rest{\MU}{d}$ of gene mutations known to be involved in disease $d$, which results from medical knowledge. In our experiments, we label magenta edges between a mutation and a disease with the so-called GDA score. This score, provided by the DisGeNET database, ranges between 0 and 1 and takes into account the number and type of sources (level of curation, model organisms), and the number of publications supporting the association between $m$ and $d$. In this setting, the set $\rest{\MU}{d}$ can be computed by considering the neighbors of $d$ in the magenta graph having GDA score close to 1.

Provided that the sample of patients is sufficiently broad, we have that $\rest{\MU}{d}$ should be contained in $\mcup{d}$: every gene mutation known to be involved in $d$ necessarily occurs in some patient with disease $d$. Next, we compare the two sets of mutations $\rest{\MU}{d}$ and $\mcap{d}$ and distinguish the following cases: if $\rest{\MU}{d}=\mcap{d}$ then the medical knowledge perfectly matches with the experimental evidence for disease $d$; otherwise, either the medical knowledge is incomplete for disease $d$ because there are gene mutations that are present in every patient with disease $d$ but are not anticipated by the current medical knowledge, or the evidence is inconsistent for disease $d$ because some patients with disease $d$ do not have a predicted gene mutation, or a combination of them. In such cases, deeper examinations are suggested.

For each $d\in Di$ computing $\rest{\MU}{d}$, $\mcup{d}$ and $\mcap{d}$ can be efficiently done through standard graph search algorithms, such as breadth-first search.
\vspace{-5mm}

\subsubsection{Partitioning patients into homogenous groups.}\label{Sec_partitioning}

Medical evidence shows ({\it e.g.}, see \cite{PPS22}) that the percentage of patients that positively react to treatments is less than expected, although drugs are chosen based on the patient specific gene mutation profile. The general feeling among experts is that concentrating on driver gene mutations is not enough.

To face this general problem, here we propose some possible approaches, all based on the idea of recognizing groups of patients that, for some reason, can be considered as similar. Then, we can propose to medical doctors a deep analysis of the gene mutations of patients in the same groups, so that they can look for the presence of specific gene mutations that the drug treatment has not targeted: some of these gene mutations could inhibit the cure and be considered responsible for the treatment failure.

As a first approach we consider a {\it similarity function} between patient genetic profiles (see, for example,~\cite{KJ19}); the idea is that if a patient has been successfully treated with a drug, patient with a similar genetic profile could be successfully administered with the same drug. More in detail, given a threshold $k$, exploiting green edges of $H$, we can determine the sets of patients such that their genetic profiles are within $k$ from each other.

This approach is related to the so-called {\it agnostic paradigm}, in which patients with very similar genetic profiles are administered with the same drug, regardless of the tumor each patient has been diagnosed. Nevertheless, the agnostic paradigm, although biologically fascinating, did not produce significant effects apart from very few cases ({\it e.g.}, NTRK \cite{MFP22}).

As an example, note that in the previous section we implicitly considered groups of patients affected by the same disease. As a second aproach, we also propose to further partition patients with the same disease $d$ according to their {\it survival period} (available from the magenta graph), clearly strongly related to the effectiveness of the administered drugs. The goal is to check if patients in the same group exhibit deeper similarities in the genetic profile with respect to the whole set of patients and provide evidence of some treatments' (in)effectiveness.
\vspace{-8mm}

\subsubsection{Optimized drug treatments.}\label{Sec_validate_treatment}

Here, we propose an algorithm joining the information obtained separately on the one hand from $G \cup M$ and on the other hand from $R$ to suggest drug treatments optimizing the benefits and minimizing adverse effects for a specific patient. 

For any gene mutation $m$, we can exploit magenta edges to deduce the set $\rest{\DR}{m}$ of the drugs that have an effect on $m$. Hypothetically, administering to a patient $p$ all the drugs in $\bigcup_{m_{p}} \rest{\DR}{m_p}$ (where $m_p$ is any gene mutation of $p$, selected through the green edges) would guarantee the best treatment for $p$. 

Nevertheless, given the possible adverse effects of these drugs (possibly depending on their interactions), only few of them can be administered simultaneously to a patient, even at the cost of ignoring some gene mutations. Indeed, in practice, only very few mutations of a patient are being treated: current drugs are designed to deal with very specific gene mutations, known as {\it target}.  Hence, given a patient $p$ and a (small) subset $Z$ of their gene mutations, the aim is to compute a drug subset $W$ of minimum size that targets all gene mutations in $Z$.  

This problem is related to the well-studied {\it minimum hitting set} problem, defined as follows. Let $U$ be a finite set and $\mathcal{U}=\{U_1,U_2,\ldots \}$ a collection of subsets of $U$. A {\it hitting set} for $\mathcal{U}$ is a subset $U'$ of $U$ such that $U_i \cap U'\neq \emptyset$, for every $i$. The {minimum hitting set problem} consists of determining a hitting set of minimum size. Computing the minimum hitting set is known to be computationally hard~\cite{SC10}.

Our problem can be modeled in terms of the minimum hitting set problem as follows: $U$ coincides with the set of the drugs $\DR$ and each $U_i$ is a $\rest{\DR}{m}$, for some $m\in Z$. Thus, solving the minimum hitting set problem on this instance gives a minimum-size drug treatment that targets all gene mutations in $Z$. 

In the literature, some papers propose similar strategies. In particular, in~\cite{Vaz09}, the authors solve the hitting set problem with a heuristic approach restricting to drug combinations of size at most three. Note that there are some papers that aim to solve the adequate drug treatment problem. The work of Johnson~\cite{Johnson73} highlights a polynomial-time heuristic approximation algorithm. Finally, in~\cite{Mezard07}, it has been developed a statistical mechanics approach to attack this problem. We point out that these previous approaches are either non-deterministic or do not obtain exact solutions.

In contrast to the previous work, we propose a deterministic and exact algorithm to solve the adequate drug treatment problem, which can be generalized by taking into account the adverse effects of drugs, minimizing both them and the number of involved drugs, thus improving the precision and safety of drug treatments. The main idea is that we add a node weight on the drug nodes related to their toxicity and solve the problem by computing a minimum weight hitting set. Given that the set of target gene mutations is small in practice, the proposed is computationally reasonable.
\vspace{-5mm}

\section{Results}\label{sec:RESULTS}

\vspace{-3mm}
In this section, we show the results of some experiments. Due to the lack of some crucial information in the public databases and the difficulty of getting some part of it, we only partially address the objectives described in Subsection 2.2; nevertheless, we try to keep the flavor of the underlying idea.

We focus on three different medical studies: Metastatic Non-Small Cell Lung Cancer~\cite{nsclc_ctdx_msk_2022} with 930 patients, MSK MetTropism~\cite{msk_impact_2021} with 24755 patients, and MSK-IMPACT Clinical Sequencing Cohort~\cite{msk_impact_2017} with 7091 patients. We chose these studies because they consider a sequencing technology guaranteeing a 500-gene panel for each patient. 
\ifshort
Nevertheless, for the sake of brevity, we report only the results concerning the first study, but the interested reader can find all the other results in \cite{vers_ArXiv}.
\fi
\vspace{-5mm}

\subsubsection{Analysis of data in public databases.}

We observed that the databases we use as reference, CBioPortal and DisGeNET, are not coherent; indeed, while the former contains specific genetic mutations, the latter deals only with mutated genes. It follows that, in order to compare the extracted results, we have to downgrade the genetic mutations to simple mutated genes. In order to have an idea about how much information we are losing in this way, we compare the data extracted from CBioPortal, counting them in different ways. In Table~\ref{tab:exp1}, we show the following information for the study MSK MetTropism:

\begin{itemize}[wide, itemsep=0.5mm, topsep=1mm]
\item the percentage of the 10 most frequent mutations with respect to the total number of mutations;
\item the percentage of the 10 most frequent mutated genes, where all the mutations on the same gene are counted;
\item the percentage of the 10 most frequent mutated genes, where multiple mutations on the same gene are counted as one.
\end{itemize}

\begin{table}[ht]
\vspace{-3mm}
\begin{center}
\begin{tabular}{|| l | c || l | c || l | c||}
\hline
gene mutations & \% & genes & \% & genes & \%\\
& & (with mult.) & & (w/o mult.) &\\
\hline
\hline
KRAS\_12\_25398284\_25398284 & 12.7 & TP53 & 43.6 & TP53 & 50.3\\
TERT\_5\_1295228\_1295228 & 6.7 & KRAS & 21.7 & KRAS & 22.0\\
KRAS\_12\_25398285\_25398285 & 4.8 & PIK3CA & 14.2 & APC & 14.7\\
PIK3CA\_3\_178936091\_178936091 & 3.3 & APC & 9.8 & PIK3CA & 14.5\\
BRAF\_7\_140453136\_140453136 & 3.2 & TERT & 9.6 & TERT & 11.5\\
PIK3CA\_3\_178952085\_178952085 & 3.2 & EGFR & 4.8 & ARID1A & 10.2\\
TP53\_17\_7578406\_7578406 & 2.7 & BRAF & 4.4 & PTEN & 7.2\\
PIK3CA\_3\_178936082\_178936082 & 2.0 & PTEN & 3.4 & KMT2D & 7.1\\
TP53\_17\_7577120\_7577120 & 1.7 & ARID1A & 3.1 & EGFR & 6.6\\
TP53\_17\_7577538\_7577538 & 1.7 & CDKN2A & 2.8 & BRAF & 6.2\\
\hline
\end{tabular}
\vspace*{3mm}
\caption{Results of an experiment performed on the 24755 patients of MSK MetTropism study: we show the 10 most frequent mutations (first columns), mutated genes with multiplicity (second columns), and mutated genes without multiplicity (third columns) together with the corresponding percentages.}\label{tab:exp1}
\end{center}
\vspace{-11mm}
\end{table}

From this data, it is evident that not only it is completely different to consider percentages of mutated genes instead of specific mutations, but it is also different to take into account multiplicity instead of ignoring it. It follows that the setup used to extract each data from the knowledge graph $H$ must be accurately detailed to medical doctors.
\vspace{-5mm}

\subsubsection{Comparing medical knowledge and data evidence: Lung Adenocarcinoma.}

We now consider only the patients characterized by the same disease $d\in \DI$. Analogously to Table~\ref{tab:exp1}, Table~\ref{tab:exp2} shows data w.r.t. the  3972 patients affected by one of the most frequent diseases included in the MSK MetTropism study, namely Lung Adenocarcinoma.

\begin{table}[t]
\begin{center}
\begin{tabular}{|| l | c || l | c || l | c||}
\hline
gene mutations & \% & genes & \% & genes & \%\\
& & (with mult.) & & (w/o mult.) &\\
\hline
\hline
KRAS\_12\_25398285\_25398285 & 14.7 & TP53 & 49.2 & TP53 & 48.5\\
KRAS\_12\_25398284\_25398284 & 13.6 & EGFR & 34.9 & KRAS & 33.9\\
EGFR\_7\_55259515\_55259515 & 9.1 & KRAS & 34 & EGFR & 29.4\\
EGFR\_7\_55242465\_55242479 & 5.0 & STK11 & 13.9 & STK11 & 16.1\\
EGFR\_7\_55242466\_55242480 & 2.9 & KEAP1 & 11.1 & KEAP1 & 14.1\\
EGFR\_7\_55249071\_55249071 & 2.9 & RBM10 & 7.7 & RBM10 & 11.7\\
U2AF1\_21\_44524456\_44524456 & 2.0 & PIK3CA & 5.5 & PTPRD & 8.8\\
PIK3CA\_3\_178936091\_178936091 & 1.8 & BRAF & 4.7 & SMARCA4 & 8.2\\
ERBB2\_17\_37880981\_37880982 & 1.6 & CDKN2A & 4.2 & ATM & 7.8\\
KRAS\_12\_25380275\_25380275 & 1.5 & SMARCA4 & 3.9 & NF1 & 7.5\\
\hline
\end{tabular}
\vspace*{3mm}
\caption{Results of an experiment performed on the MSK MetTropism study on the 3972 patients affected by lung adenocarcinoma:
we show the 10 most frequent mutations (first columns), mutated genes with multiplicity (second columns), and mutated genes without multiplicity (third columns) together with the corresponding percentages.}\label{tab:exp2}
\end{center}
\vspace*{-15mm}
\end{table}

Comparing Tables~\ref{tab:exp1} and~\ref{tab:exp2}, one can observe sensible differences: for example, the gene mutation KRAS\_12\_25398285\_25398285 appears in only 4.8\% of all patients while in 14.7\% of those affected by Lung Adenocarcinoma. Moreover, the gene mutation TERT\_5\_1295228\_1295228 appears in 6.7\% of the patients in Table~\ref{tab:exp1} while is negligible in Table~\ref{tab:exp2}. An even more notable discrepancy can be observed in the gene EGFR: only 6.6\% of the total population of patients has this gene mutated, while the percentage increases to 29.4 for the patients affected by Lung Adenocarcinoma. These considerations are not meant to infer any conclusion at the medical level but, especially if joined with similar studies, aim to suggest a direction for further research.

\begin{table}[ht]
\vspace{-3mm}
\begin{center}
\begin{tabular}{|| c | c || c | c || c | c ||}
\hline
mutated gene & GDA Score & mutated gene & GDA Score & mutated gene & GDA Score\\
\hline
\hline
BRAF & 1.0 & FGFR2 & 0.85 & MAP2K1 & 0.8\\
ALK & 1.0 & AKT1 & 0.85 & CTNNB1 & 0.8\\
ROS1 & 1.0 & MUC5AC & 0.85 & CDKN2A & 0.8\\
KRAS & 1.0 & TYMS & 0.85 & RAF1 & 0.8\\
EGFR & 1.0 & CCND1 & 0.85 & CHRNA3 & 0.8\\
ERBB2 & 0.95 & STK11 & 0.8 & FGFR3 & 0.8\\
PIK3CA & 0.95 & TERT & 0.8 & ATM & 0.8\\
TP53 & 0.95 & FGFR4 & 0.8 & EGF & 0.8\\
MYC & 0.9 & HRAS & 0.8 & &\\
\hline
\end{tabular}
\vspace{3mm}
\caption{Mutated genes with GDA score at least 0.8 in lung adenocarcinoma.}\label{tab:gda}
\end{center}
\vspace{-12mm}
\end{table}

On the one hand, it is clear that no single gene mutation appears in all patients affected by lung adenocarcinoma. Therefore, $\mcap{d}$ is trivially empty. On the other hand six of the genes appearing in Table~\ref{tab:gda}, namely BRAF, KRAS, EGFR, STK11, ATM, and TP53, are also represented in  Table~\ref{tab:exp2} showing a level of agreement between medical knowledge on lung adenocarcinoma disease and the evidence collected on the patients. Anyway, some genes appearing in Table~\ref{tab:gda} with GDA score 1, namely ALK and ROS1, do not appear in Table~\ref{tab:exp2} and hence in $\mcup{d}$, and some genes appearing in Table~\ref{tab:exp2} with frequency above 10\% without multiplicity, namely KEAP1 and RBM10, do not appear in Table~\ref{tab:gda}, showing some inconsistencies between medical knowledge and data that we have considered.
\vspace{-5mm}

\subsubsection{Partitioning patients into homogenous groups: survival period.}

Estimating the effectiveness of drug treatment is a difficult task because it takes into account different parameters. One of these parameters is the survival period.

We partition the patient population into three sets: the first one $\PA_{\geq 36}=\{p\in \PA~|~\rho(p)\geq 36\}$ contains all the patients whose survival period is of at least 36 months, the second one $\PA_{\leq 6}=\{p\in \PA~|~\rho(p)\leq 6\wedge \alpha(p)=F\}$ contains all the patients whose survival period is of at most 6 months and are marked as deceased, and the third set contains all the remaining patients. 

\begin{table}
\begin{center}
\begin{tabular}{|| l | c || l | c || l | c||}
\hline
gene mutations & \% & genes & \% & genes & \%\\
& & (with mult.) & & (w/o mult.) &\\
\hline
\hline
KRAS\_12\_25398284\_25398284 & 8.6 & TP53 & 35.6 & TP53 & 38.8\\
TERT\_5\_1295228\_1295228 & 6.6 & PIK3CA & 16.9 & PIK3CA & 17\\
PIK3CA\_3\_178952085\_178952085 & 4.7 & KRAS & 15.4 & KRAS & 15.7\\
KRAS\_12\_25398285\_25398285 & 3.6 & TERT & 9.9 & APC & 13.7\\
PIK3CA\_3\_178936091\_178936091 & 3.5 & APC & 9.7 & TERT & 11.5\\
BRAF\_7\_140453136\_140453136 & 3.0 & EGFR & 6.6 & ARID1A & 9.3\\
PIK3CA\_3\_178936082\_178936082 & 2.6 & BRAF & 4.8 & EGFR & 7.8\\
TP53\_17\_7578406\_7578406 & 2.1 & PTEN & 4.7 & PTEN & 7.2\\
EGFR\_7\_55259515\_55259515 & 1.8 & ARID1A & 3.7 & FAT1 & 6.2\\
TP53\_17\_7577538\_7577538 & 1.4 & CTNNB1 & 3.0 & PTPRT & 6.1\\
\hline
\end{tabular}
\vspace*{3mm}
\caption{Results of the experiment on the MSK MetTropism study on the 5295 patients with a survival period of at least 36 months:
we show the 10 most frequent mutations (first columns), mutated genes with multiplicity (second columns), and mutated genes without multiplicity (third columns) together with the corresponding percentages.}\label{tab:exp3}
\end{center}
\vspace{-1mm}
\end{table}

\begin{table}[h]
\vspace*{-1cm}
\begin{center}
\begin{tabular}{|| l | c || l | c || l | c||}
\hline
gene mutations & \% & genes & \% & genes & \%\\
& & (with mult.) & & (w/o mult.) &\\
\hline
\hline
KRAS\_12\_25398284\_25398284 & 15.3 & TP53 & 57.8 & TP53 & 62.1\\
TERT\_5\_1295228\_1295228 & 8.2 & KRAS & 27.9 & KRAS & 27.9\\
KRAS\_12\_25398285\_25398285 & 7.1 & TERT & 12.1 & TERT & 13.3\\
BRAF\_7\_140453136\_140453136 & 3.3 & PIK3CA & 11.4 & PIK3CA & 12.2\\
PIK3CA\_3\_178952085\_178952085 & 2.8 & APC & 6.4 & ARID1A & 10.2\\
TP53\_17\_7578406\_7578406 & 2.7 & CDKN2A & 6.0 & APC & 10.1\\
PIK3CA\_3\_178936091\_178936091 & 2.6 & BRAF & 5.2 & CDKN2A & 9.5\\
PIK3CA\_3\_178936082\_178936082 & 2.3 & STK11 & 4.0 & KEAP1 & 7.2\\
TP53\_17\_7577538\_7577538 & 2.2 & EGFR & 3.7 & STK11 & 6.9\\
TP53\_17\_7577094\_7577094 & 2.0 & SMAD4 & 3.3 & RB1 & 6.8\\
\hline
\end{tabular}
\vspace*{3mm}
\caption{Results of the experiment on the MSK MetTropism study on the 2768 patients that have a survival period of at most 6 months: we show the 10 most frequent mutations (first column), mutated genes with multiplicity (second column), and mutated genes without multiplicity (third column) together with the corresponding percentages.
}\label{tab:exp4}
\end{center}
\vspace{-8mm}
\end{table}

We selected the 5295 patients of the MSK MetTropism study in $\PA_{\geq 36}$ and the 2768 patients in $\PA_{\leq 6}$ and summarized the results in Tables~\ref{tab:exp3} and~\ref{tab:exp4}, respectively. Comparing Table~\ref{tab:exp1} with Tables~\ref{tab:exp3} and~\ref{tab:exp4}, one can observe that the distribution of the percentages of their mutations completely changes. As an example, TP53, KRAS, and TERT dramatically increase their percentages, whereas PIK3CA, EGFR, and STR11 decrease significantly their percentages. This behavior can be explained by the medical awareness that certain combinations of mutations indicate either a different response to treatments or a different evolution of the disease. A deep study of these results should be performed by medical doctors, who could individuate interesting combinations of gene mutations, both in the population of patients with a low survival period and in that one with a high survival period.
\vspace{-4mm}

\subsubsection{Partitioning patients into homogenous groups: genetic mutation profile.}

Moreover, to understand whether there are some groups of patients that can be considered similar, we implemented two different similarity measures based on the genetic profiles. The {\it Hamming distance}~\cite{ham50} between two patients counts the number of gene mutations that affect only one of them. The {\it Jaccard distance}~\cite{jaccard1912distribution} is a variation of the Hamming distance where the value is normalized by the total number of gene mutations affecting the two considered patients, taking into account the inequalities due to the possibly imbalanced number of observed gene mutations or different gene sequencing ({\it e.g.}, different number of checked genes). Clearly, two patients having a similar genetic profile are also very close with respect to the considered measures.

The overall idea is that patients who are grouped together, whether they have either Hamming or Jaccard distance small, might experience the same disease, similar disease evolution, and comparable responses to drug treatments. 

Regardless of the similarity measure used, our experiments show that most of the patients are isolated, that is, the groups of similar patients are very often singletons. As an example, we report the results obtained from the MSK MetTropism study when considering patients at Hamming distance at most 10 from each other. Out of 24755 patients, there are only 6 groups that are not singletons which include a total of 14 patients. This means, at least considering the data at our disposal, that it is very unlikely that any two patients are similar from the genetic profile point of view.

As expected, these results confirm that genetic similarities are not enough to explain different behaviors of the human body with respect to oncology medicine. Since the techniques we use to aggregate patients are not the most sophisticated nor the most appropriate for this specific task, in the following we propose more advanced clustering techniques.
\vspace{-5mm}

\subsubsection{Paritioning Patient in homogenous groups: coexisting mutations.}

\begin{table}[t]
\begin{center}
\begin{tabular}{|| l | c | c | c ||}
\hline
gene mutations & \% patients & \% living patients & \% deceased patients\\
\hline
\hline
KRAS & 100 & 61.3 & 38.7\\
EGFR & 100 & 61.3 & 38.7\\
TP53 & 52.7 & 51.0 & 49.0\\
APC & 45.2 & 76.2 & 23.9\\
ARID1A & 38.7 & 83.3 & 16.7\\
KMT2D & 35.5 & 93.9 & 6.1 \\
PIK3CA & 35.5 & 90.9 & 9.1\\
FAT1 & 35.5 & 81.8 & 18.2\\
ATM & 33.3 & 77.4 & 22.6\\
PTEN & 31.2 & 89.7 & 10.3\\
\hline
\end{tabular}
\vspace*{3mm}
\caption{Results of an experiment performed on the MSK MetTropism study on the 93 patients that have both EGFR and KRAS genes mutated: we show the 10 most frequent mutations (first column), the percentage of patients with that gene mutated (second column), and the percentages of living and deceased patients among those with that gene mutated (third and fourth columns).}\label{tab_EGFR_KRAS}
\end{center}
\vspace{-10mm}
\end{table}

We wonder whether there exist some combinations of gene mutations that appear simultaneously in significant portions of the patient population: we compute all the (maximal) $k$-coexisting-mutation sets, {\it i.e.}, sets of mutations simultaneously present in at least $k\%$ of patients, and return these sets of patients. 

From the evaluation of the data, we observe that $k$-coexisting-mutation sets are made of a single mutation, even for very small values of $k$. Considering the MSK MetTropism study again, there is only one $k$-coexisting-mutation set, with $k=12$, which consists of the single mutation KRAS\_12\_25398284\_25398284. It seems unrealistic to assume that every patient affected by KRAS\_12\_25398284\_ 25398284 can be considered similar with respect to the affected disease, disease evolution, and response to drug treatments.

Some combinations of gene mutations are particularly relevant for medical doctors: contemporary mutation in genes EGFR and KRAS is one of them. So, we extracted all the patients with both these two genes mutated, whose 61.3\% of them were alive at the time of the study. It is natural to wonder whether there is an explanation for the alive patients to survive. For each analyzed gene, we computed the fraction (expressed as a percentage) of the patients having that gene mutated in three different patient populations: all patients, the living, and the deceased ones. It turns out that the patients with certain further mutations (such as ARID1A, PIK3CA, AT1, or PTEN) are much more likely to survive, as shown in Table \ref{tab_EGFR_KRAS}, indeed the percentage of living patients is more than 80\% in the presence of these gene mutations. This kind of table could be of interest to better understand whether there are special combinations of gene mutations that significantly increase the survival probability. The results of this experiment for other combinations of genes, such as EGFR and T790R, do not provide the same interesting output: for example, in all three considered studies, no patient had these two genes mutated at the same time.
\vspace{-4mm}

\section{Conclusions}\label{sec:CONCLUSIONS}
\vspace{-2mm}

In this paper, we designed a unified graph-based representation of medical data for precision oncology medicine and proposed three possible applications whose solutions exploit known results from theoretical computer science.
 
Our approach's novelty lies in how we store and deduce information. In particular: 
\begin{itemize}[wide, itemsep=-0.5mm, topsep=0.5mm]
\item we develop a knowledge graph that exploits various databases  to deduce fundamental information using graph-theoretic tools;
\item we implement a deterministic framework to infer personalized medical information in contrast to past research strategies that have been using data aggregation~\cite{CV19,Mal18,Ual20}, pattern recognition~\cite{CZ17,Hal24} and statistical performance~\cite{I23,Val21};
\item our knowledge graph model allows one for quick and efficient updates, whether there is a new node or some information has changed, in contrast to static models based on machine learning techniques (see for example~\cite{Gal21}).
\end{itemize}
\vspace{-2mm}

\begin{credits}
\subsubsection{\ackname}
The authors would like to thank medical doctors Gennaro Daniele and Pasquale Lombardi for the exciting discussions on cancer handling from a medical point of view, Kilian Schulz for contributing to extracting data from databases during his honors program, Luciano Giac\`o and Federica Persiani for their helpful feedback on the databases to be used.

This work was supported by {\it Sapienza} University of Rome, project title: {\it Graph models for precision oncology medicine}, grant number: RM122181612C08BB.

\iflong
This paper was published as an extended abstract in the proceedings of CIBB 2024, 19th conference on Computational Intelligence methods for Bioinformatics and Biostatistics.
\fi
\end{credits}

\bibliographystyle{splncs04}
\bibliography{references}

\iflong
\newpage
\appendix
\section*{Appendix}
\section{Tables of the results on the other studies}

Tables~\ref{tab:a} and~\ref{tab:n} correspond to Table \ref{tab:exp1} for the other two studies.

\begin{table}[h!]
\begin{center}
\begin{tabular}{|| l | c || l | c || c | c||}
\hline
gene mutations & \% & genes & \% & genes & \%\\
& & (with mult.) & & (w/o mult.) &\\
\hline
\hline
KRAS\_12\_25398285\_25398285 & 11.9 & TP53 & 73.7 & TP53 & 62.4\\
EGFR\_7\_55259515\_55259515 & 11.5 & EGFR & 45.2 & EGFR & 32.6\\
KRAS\_12\_25398284\_25398284 & 11.5 & KRAS & 28.0 & KRAS & 26.9\\
EGFR\_7\_55242465\_55242479 & 6.7 & KEAP1 & 12.4 & KEAP1 & 12.2\\
EGFR\_7\_55249071\_55249071 & 5.2 & STK11 & 11.5 & STK11 & 11.4\\
TP53\_17\_7577120\_7577120 & 3.2 & MET & 10.8 & MET & 8.7\\
TP53\_17\_7578406\_7578406 & 2.5 & ERBB2 & 7.5 & RBM10 & 6.9\\
EGFR\_7\_55242466\_55242480 & 2.3 & PIK3CA & 6.9 & PIK3CA & 6.7\\
PIK3CA\_3\_178936091\_178936091 & 1.7 & RBM10 & 6.9 & SMARCA4 & 6.1\\
TP53\_17\_7577121\_7577121 & 1.7 & PTPRD & 6.3 & ERBB2 & 6.1\\
\hline
\end{tabular}
\vspace*{5mm}
\caption{Results of an experiment performed on the 930 patients of Metastatic Non-Small Cell Lung Cancer study.}\label{tab:a}
\end{center}
\end{table}
\vspace*{-1cm}

\begin{table}[h!]
\begin{center}
\begin{tabular}{|| l | c || l | c || c | c||}
\hline
gene mutations & \% & genes & \% & genes & \%\\
& & (with mult.) & & (w/o mult.) &\\
\hline
\hline
\footnotesize TERT\_5\_1295228\_1295228 & 9.8 & TP53 & 42.3 & TP53 & 46.9\\
KRAS\_12\_25398284\_25398284 & 9.4 & KRAS & 17.2 & KRAS & 17.5\\
KRAS\_12\_25398285\_25398285 & 4.3 & TERT & 13.7 & TERT & 14.9\\
PIK3CA\_3\_178936091\_178936091	& 3.2 & PIK3CA & 12.8 & PIK3CA & 13.4\\
PIK3CA\_3\_178952085\_178952085	& 3.2 & APC & 6.3 & APC & 10.9\\
BRAF\_7\_140453136\_140453136 & 3.1 & EGFR & 6.0 & MLL2 & 9.3\\
TERT\_5\_1295250\_1295250 & 3.1 & BRAF & 4.8 & ARID1A & 9.2\\
PIK3CA\_3\_178936082\_178936082	& 2.2 & PTEN & 3.0	 & EGFR & 7.3\\
TP53\_17\_7578406\_7578406 & 2.1 & CDKN2A & 2.7 & PTEN & 7.0\\
TP53\_17\_7577120\_7577120 & 1.7 & CTNNB1 & 2.5 & MLL3 & 6.9\\
\hline
\end{tabular}
\vspace*{5mm}
\caption{Results of an experiment performed on the 7091 patients of MSK-IMPACT Clinical Sequencing Cohort study.}\label{tab:n}
\end{center}
\end{table}

\newpage
Tables~\ref{tab:d} and~\ref{tab:q} correspond to Table \ref{tab:exp2} for the other two studies.
\vspace*{-5mm}

\begin{table}[ht]
\begin{center}
\begin{tabular}{|| l | c || l | c || l | c||}
\hline
gene mutations & \% & genes & \% & genes & \%\\
& & (with mult.) & & (w/o mult.) &\\
\hline
\hline
KRAS\_12\_25398285\_25398285 & 13.0 & TP53 & 67.8 & TP53 & 60.5\\
EGFR\_7\_55259515\_55259515 & 13.0 & EGFR & 49.7 & EGFR & 36.2\\
KRAS\_12\_25398284\_25398284 & 11.3 & KRAS & 29.3 & KRAS & 28.3\\
EGFR\_7\_55242465\_55242479 & 7.4 & KEAP1 & 11.5 & KEAP1 & 13.0\\
EGFR\_7\_55249071\_55249071 & 6.0 & STK11 & 10.4 & STK11 & 11.5\\
TP53\_17\_7577120\_7577120 & 3.1 & MET & 9.8 & MET & 8.6\\
EGFR\_7\_55242466\_55242480 & 2.6 & ERBB2 & 7.7 & RBM10 & 7.2\\
TP53\_17\_7578406\_7578406 & 2.4 & RBM10 & 6.3 & PIK3CA & 6.5\\
TP53\_17\_7577538\_7577538 & 1.9 & PIK3CA & 6.2 & ERBB2 & 6.5\\
TP53\_17\_7577121\_7577121 & 1.7 & PTPRD & 5.4 & PTPRD & 6.0\\
\hline
\end{tabular}
\vspace*{3mm}
\caption{Results of an experiment performed on the 802 patients of Metastatic Non-Small Cell Lung Cancer study affected by lung adenocarcinoma.}\label{tab:d}
\end{center}
\end{table}
\vspace*{-1cm}

\begin{table}[ht]
\begin{center}
\begin{tabular}{|| l | c || l | c || l | c||}
\hline
gene mutations & \% & genes & \% & genes & \%\\
& & (with mult.) & & (w/o mult.) &\\
\hline
\hline
KRAS\_12\_25398285\_25398285 & 14.4	 & TP53 & 61 & TP53 & 53.2\\
KRAS\_12\_25398284\_25398284 & 12.5 & EGFR & 37.9 & KRAS & 31.9\\
EGFR\_7\_55259515\_55259515	 & 8.8 & KRAS & 33 & EGFR & 28.3\\
EGFR\_7\_55242465\_55242479 & 5.0 & STK11 & 21.7 & STK11 & 18.5\\
EGFR\_7\_55249071\_55249071 & 4.4 & KEAP1 & 20.5 & KEAP1 & 16.3\\
EGFR\_7\_55242466\_55242480 & 3.3 & PTPRD & 14.6	 & PTPRD & 9.9\\
PIK3CA\_3\_178936091\_178936091 & 2.2 & SMARCA4 & 13.0 & SMARCA4 & 9.8\\
U2AF1\_21\_44524456\_44524456 & 1.9 & RBM10 & 12.8 & RBM10 & 9.8\\
BRAF\_7\_140453136\_140453136 & 1.9 & PTPRT & 11.3 & FAT1 & 8.2\\
TP53\_17\_7577120\_7577120 & 1.5 & ATM & 11.1 & ARID1A & 7.8\\
\hline
\end{tabular}
\vspace*{3mm}
\caption{Results of an experiment performed on the 1239 patients of MSK-IMPACT Clinical Sequencing Cohort study affected by lung adenocarcinoma.}\label{tab:q}
\end{center}
\end{table}

\newpage
Tables~\ref{tab:o},~\ref{tab:c},~\ref{tab:p} and~\ref{tab:b} correspond to Tables \ref{tab:exp3} and \ref{tab:exp4} for the other two studies.
\vspace*{-5mm}

\begin{table}[ht]
\begin{center}
\begin{tabular}{|| l | c || l | c || l | c||}
\hline
gene mutations & \% & genes & \% & genes & \%\\
& & (with mult.) & & (w/o mult.) &\\
\hline
\hline
TERT\_5\_1295228\_1295228 & 11.8	 & TP53 & 37.8 & TP53 & 35.3\\
TP53\_17\_7578406\_7578406 & 5.9	 & APC & 26.1 & TERT & 14.7\\
KRAS\_12\_25398284\_25398284 & 5.9 & MAP3K1 & 20.3 & MAP3K1 & 11.8\\
BRAF\_7\_140453155\_140453155 & 2.9 & TERT & 17.6 & MLL3 & 11.8\\
RFWD2\_1\_176054935\_176054935 & 2.9 & MLL3 & 17.4 & KRAS & 11.8\\
FAT1\_4\_187584651\_187584651 & 2.9 & PTEN & 14.5 & SETD2 & 8.8\\
APC\_5\_112174059\_112174059 & 2.9 & KRAS & 11.7 & FGFR2 & 8.8\\
APC\_5\_112174426\_112174426 & 2.9 & ARID1A & 11.6 & APC & 8.8\\
APC\_5\_112174432\_112174432 & 2.9 & FAT1 & 8.7 & PTEN & 8.8\\
APC\_5\_112178449\_112178449 & 2.9 & SETD2 & 8.7 & TSC2 & 8.8\\
\hline
\end{tabular}
\vspace*{3mm}
\caption{Results of an experiment performed on the 34 patients of MSK-IMPACT Clinical Sequencing Cohort study with a survival period of at least 36 months.}\label{tab:o}
\end{center}
\end{table}

\vspace*{-1.3cm}

\begin{table}[ht]
\begin{center}
\begin{tabular}{|| l | c || l | c || l | c||}
\hline
gene mutations & \% & genes & \% & genes & \%\\
& & (with mult.) & & (w/o mult.) &\\
\hline
\hline
EGFR\_7\_55249071\_55249071 & 14.2 & TP53 & 78.5 & TP53 & 58.5\\
EGFR\_7\_55259515\_55259515 & 14.2 & EGFR & 74 & EGFR & 42.6\\
KRAS\_12\_25398285\_25398285 & 13.1 & KRAS & 24.1 & KRAS & 21.6\\
EGFR\_7\_55242465\_55242479 & 12.5 & MET & 15.9 & MET & 10.2\\
KRAS\_12\_25398284\_25398284 & 7.4 & KMT2D & 12.6 & PIK3CA & 7.4\\
TP53\_17\_7577538\_7577538 & 3.4 & ARID1A & 10.1 & ALK & 7.4\\
CTNNB1\_3\_41266113\_41266113 & 2.8 & ALK & 10.1 & ARID1A & 6.8\\
EGFR\_7\_55249092\_55249092 & 2.8 & FAT1 & 9.6 & RBM10 & 6.8\\
EGFR\_7\_55249091\_55249091 & 2.8 & ERBB2 & 9.5 & ZFHX3 & 6.8\\
BRAF\_7\_140453136\_140453136 & 2.3 & SETD2 & 9.0 & SETD2 & 6.8\\
\hline
\end{tabular}
\vspace*{3mm}
\caption{Results of an experiment performed on the 176 patients of Metastatic Non-Small Cell Lung Cancer study with a survival period of at least 36 months.}\label{tab:c}
\end{center}
\end{table}

\newpage

\begin{table}[ht]
\begin{center}
\begin{tabular}{|| l | c || l | c || l | c||}
\hline
gene mutations & \% & genes & \% & genes & \%\\
& & (with mult.) & & (w/o mult.) &\\
\hline
\hline
TERT\_5\_1295228\_1295228 & 9.7 & TP53 & 54.9	 & TP53 & 56.2\\
KRAS\_12\_25398284\_25398284 & 8.9 & KRAS & 21.1	 & KRAS & 21.2\\
KRAS\_12\_25398285\_25398285 & 7.8 & TERT & 16 & TERT & 16.1\\
BRAF\_7\_140453136\_140453136 & 4.2	 & PIK3CA & 11.7	 & STK11 & 12.5\\
TERT\_5\_1295250\_1295250 & 3.8	 & STK11 & 10.4 & KEAP1 & 12\\
PIK3CA\_3\_178936082\_178936082 & 3.5 & KEAP1 & 9.7 & PIK3CA & 11.3\\
PIK3CA\_3\_178952085\_178952085 & 2.9 & MLL2 & 9.5 & MLL2 & 10.7\\
PIK3CA\_3\_178936091\_178936091 & 2.2 & ARID1A & 9.3 & ARID1A & 10.3\\
TP53\_17\_7577120\_7577120 & 2.2 & BRAF & 7.6 & SMARCA4 & 8.3\\
NRAS\_1\_115256529\_115256529 & 1.7 & APC & 7.6 & BRAF & 8.1\\
\hline
\end{tabular}
\vspace*{2mm}
\caption{Results of an experiment performed on the 689 patients of MSK-IMPACT Clinical Sequencing Cohort study with a survival period of at most 6 months and are deceased.}\label{tab:p}
\end{center}
\end{table}

\begin{table}[ht]
\begin{center}
\begin{tabular}{|| l | c || l | c || l | c||}
\hline
gene mutations & \% & genes & \% & genes & \%\\
& & (with mult.) & & (w/o mult.) &\\
\hline
\hline
KRAS\_12\_25398285\_25398285 & 19.7 & TP53 & 82.6 & TP53 & 64.8\\
KRAS\_12\_25398284\_25398284 & 14.1 & KRAS & 43.6 & KRAS & 42.3\\
TP53\_17\_7577120\_7577120 & 4.2 & KEAP1 & 27.3 & KEAP1 & 26.1\\
TP53\_17\_7578406\_7578406 & 3.5 & STK11 & 23.8 & STK11 & 22.5\\
TP53\_17\_7577538\_7577538	 & 2.8 & EGFR & 14.0 & EGFR & 13.4\\
TP53\_17\_7579311\_7579311 & 2.8 & APC & 9.8 & SMARCA4 & 9.2\\
KRAS\_12\_25398281\_25398281 & 2.8 & SMARCA4 & 9.1 & RBM10 & 8.5\\
TERT\_5\_1295228\_1295228 & 2.1 & PIK3CA & 8.4 & APC & 7.7\\
PIK3CA\_3\_178936082\_178936082 & 2.1 & RBM10 & 8.4 & NF1 & 7.0\\
KRAS\_12\_25398284\_25398285 & 2.1 & ATM & 8.4 & ATM & 7.0\\
\hline
\end{tabular}
\vspace*{5mm}
\caption{Results of an experiment performed on the 142 patients of Metastatic Non-Small Cell Lung Cancer study with a survival period of at most 6 months and deceased.}\label{tab:b}
\end{center}
\end{table}
\fi

\end{document}